\documentclass[10pt,twocolumn,letterpaper]{article}

\usepackage{iccv}
\usepackage{times}
\usepackage{epsfig}
\usepackage{graphicx}
\usepackage{amsmath}
\usepackage{amssymb}

\usepackage{url}
\usepackage{multirow}
\usepackage{booktabs} % for professional tables
\usepackage{tikz}
\usepackage{amsfonts}
\usepackage{nicefrac}
\usepackage{siunitx}
\usepackage{textcomp}
\usepackage{url}
\usepackage{color}
\usepackage{xcolor,colortbl}
\usepackage{xspace}
\usepackage{textcomp}
\usepackage{lipsum}
\usepackage{pifont}% http://ctan.org/pkg/pifont
\usepackage{multirow}
\usepackage{aliascnt}
\usepackage{threeparttable}
\usepackage{wrapfig}
\usepackage{soul}
\usepackage{algorithm}
\usepackage{algpseudocode}
\usepackage{tikz}
\usepackage{caption}
\usepackage{subcaption}
\usepackage{makecell}
\usepackage{wrapfig}
\usepackage{xcolor}
\usepackage{todonotes}
\usepackage{soul}
\usepackage[accsupp]{axessibility}

\usepackage{tikz}

\newcommand{\coop}{CoOp } % need the space
\newcommand{\cocoop}{CoCoOp } % need the space

\definecolor{Gray}{gray}{0.9}
\definecolor{palegray}{HTML}{F5F5F5}
\definecolor{barred}{HTML}{F24B4B}
\definecolor{barblue}{HTML}{1DB6F2}
\definecolor{baryellow}{HTML}{F2A922}
\definecolor{barpurpule}{HTML}{A441BF}
\definecolor{bargreen}{HTML}{97BF5A}
\definecolor{palepurpule}{HTML}{FFF8FF}
\definecolor{palegreen}{HTML}{F6FAF3}

\newcolumntype{b}{>{\columncolor{palegreen}}c}

\newcommand*\rot{\rotatebox{90 }}

\newcommand{\sqboxs}{1.5ex}% the square size
\newcommand{\sqbox}[1]{\textcolor{#1}{\rule{\sqboxs}{\sqboxs}}}

\newcommand{\x}{\mathbf{x}}

\newcommand{\txt}{\mathbf{t}}

\newcommand{\w}{\mathbf{w}}
\newcommand{\e}{\mathbf{e}}
\newcommand{\p}{\mathbf{p}}

\def\rvr{{\mathbf{r}}}

\definecolor{Gray}{gray}{0.9}
\definecolor{palegray}{HTML}{F5F5F5}
\definecolor{barred}{HTML}{F24B4B}
\definecolor{barblue}{HTML}{1DB6F2}
\definecolor{baryellow}{HTML}{F2A922}
\definecolor{barpurpule}{HTML}{A441BF}
\definecolor{bargreen}{HTML}{97BF5A}
\definecolor{palepurpule}{HTML}{FFF8FF}
\definecolor{palegreen}{HTML}{F6FAF3}

\newcolumntype{b}{>{\columncolor{palegreen}}c}

\def\rvr{{\mathbf{r}}}

\usepackage[breaklinks=true,bookmarks=false]{hyperref}

\iccvfinalcopy % *** Uncomment this line for the final submission

\def\iccvPaperID{****} % *** Enter the ICCV Paper ID here

\ificcvfinal\fi

\begin{document}

%%%%%%%%% TITLE
\title{\st{Variational prompt learning improves generalization of vision-language models}}
%CS: alternative suggestion
\title{Bayesian Prompt Learning for Image-Language Model Generalization}
% \title{Bayesian Prompt Learning for Vision-Language Model Transfer Learning}

\author{Mohammad Mahdi Derakhshani\textsuperscript{\rm 1}\thanks{Most work done during an internship at Samsung AI Cambridge. %Corresponding author: Mohammad Mahdi Derakhshani; Email: \texttt{m.m.derakhshani@uva.nl}
},  Enrique Sanchez\textsuperscript{\rm 5}, Adrian Bulat\textsuperscript{\rm 3, 5}\\ Victor Guilherme Turrisi da Costa\textsuperscript{\rm 2}\footnotemark[1], 
Cees G. M. Snoek\textsuperscript{\rm 1}\thanks{Equal advising}, Georgios Tzimiropoulos\textsuperscript{\rm 4,5}\footnotemark[2], Brais Martinez\textsuperscript{\rm 5}\footnotemark[2]\\
\textsuperscript{\rm 1}University of Amsterdam \hspace{2mm} \textsuperscript{\rm 2}University of Trento\hspace{2mm} 
\textsuperscript{\rm 3}Technical University of Iasi \\
\textsuperscript{\rm 4}Queen Mary University of London \hspace{2mm} 
\textsuperscript{\rm 5}Samsung AI Cambrdige\\
% \href{http://www.overleaf.com}{Something Linky}\\
%{\tt\small \{j.zhao3, cgmsnoek\}@uva.nl; \{yz593, marsic\}@rutgers.edu;}\\ \tt\small {\{xxnl,hxen,bshuai,kaustavk,xumingze,chunhliu,yuanjx,dmodolo,tighej\}@amazon.com}
}

% \author{Mohammad Mahdi Derakhshani\\
% University of Amsterdam
% \and
% Enrique Sanchez\\
% Samsung AI Cambridge
% \\
% \and
% Adrian Bulat\\
% Samsung AI Cambridge
% \\
% \and
% Victor Guilherme Turrisi da Costa\\
% University of Trento
% \and
% Cees G. M. Snoek\\
% University of Amsterdam
% \and
% Georgios Tzimiropoulos\\
% Queen Mary University London\\
% Samsung AI Cambridge
% \and
% Brais Martinez\\
% Samsung AI Cambridge
% }

\maketitle
\ificcvfinal\thispagestyle{empty}\fi

%%%%%%%%% ABSTRACT
\begin{abstract}
Foundational image-language models have generated considerable interest due to their efficient adaptation to downstream tasks by prompt learning. Prompt learning treats part of the language model input as trainable while freezing the rest, and optimizes an Empirical Risk Minimization objective. However, Empirical Risk Minimization is known to suffer from distributional shifts which hurt generalizability to prompts unseen during training. By leveraging the regularization ability of Bayesian methods, we frame prompt learning from the Bayesian perspective and formulate it as a variational inference problem. Our approach regularizes the prompt space, reduces overfitting to the seen prompts and improves the prompt generalization on unseen prompts. Our framework is implemented by modeling the input prompt space in a probabilistic manner, as an a priori distribution which makes our proposal compatible with prompt learning approaches that are unconditional or conditional on the image. We demonstrate empirically on 15 benchmarks that Bayesian prompt learning provides an appropriate coverage of the prompt space, prevents learning spurious features, and exploits transferable invariant features. This results in better generalization of unseen prompts, even across different datasets and domains.

Code available at: \color{barpurpule}{https://github.com/saic-fi/Bayesian-Prompt-Learning}
\end{abstract}

% for paper final submission
\begin{figure}[t!]
\centering
\includegraphics[width=0.9\linewidth]{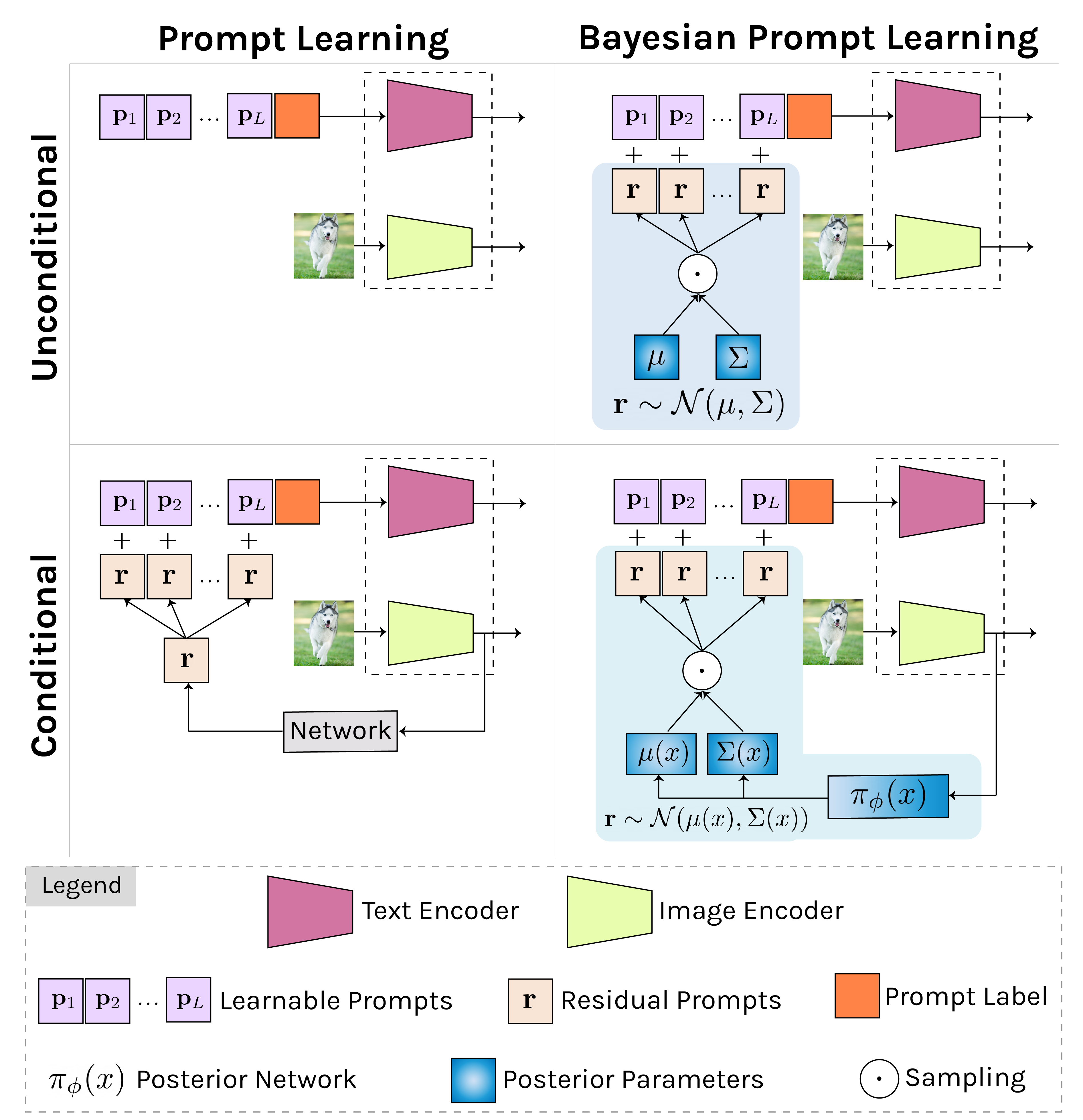}
\vspace{-2mm}
\caption{
 We present a Bayesian perspective on prompt learning by formulating it as a variational inference problem (right column). Our framework  models the prompt space as an a priori distribution which makes our proposal compatible with common prompt learning approaches that are unconditional (top) or conditional on the image (bottom).}
\vspace{-5mm}
\label{fig:splash}
\end{figure}

%%%%%%%%% BODY TEXT
\vspace{-1em}
\section{Introduction}
\vspace{-0.5em}
\label{seq:intro}

In the continuous quest for better pre-training strategies, models based on image and language supervision have set impressive milestones, with CLIP~\cite{clip_icml21}, ALIGN~\cite{align_icml21} and Flamingo~\cite{flamingo_arxiv22} being leading examples. Contrastively trained image-language models consist of image and text encoders that align semantically-related concepts in a joint embedding space. Such models offer impressive zero-shot image classification by using the text encoder to generate classifier weights from arbitrarily newly defined category classes without relying on any visual data. In particular, the class name is used within a handcrafted \textit{prompt} template and then tokenized and encoded into the shared embedding space to generate new classifier weights. Rather than manually defining prompts, both Lester \etal \cite{lester2021power} and Zhou \etal \cite{zhou2022learning} demonstrated prompts can instead be optimized in a data-driven manner through backpropagation. However, as prompt learning typically has access to only a few training examples per prompt, overfitting to the seen prompts in lieu of the unseen prompts is common \cite{zhou2022learning}. In this paper, we strive to mitigate the overfitting behavior of prompt learning so as to improve generalization for unseen prompts.

% for paper review.
% \begin{figure}[t!]
% \centering
% \includegraphics[width=0.95\linewidth]{images/Figure1_karla_2.png}
% \vspace{-2mm}
% \caption{
%  We frame prompt learning from the Bayesian perspective and formulate it as a variational inference problem (right column). Our framework is implemented by modeling the input prompt space in a probabilistic manner, as an a priori distribution which makes our proposal compatible with common prompt learning approaches that are unconditional (top) or conditional on the image (bottom).}
% \vspace{-5mm}
% \label{fig:splash}
% \end{figure}

Others before us have considered the generalization problem in prompt learning as well, \eg,~\cite{zhou2022learning, zhou2022conditional}, be it they all seek to optimize a so-called Empirical Risk Minimization. It is, however, well known that Empirical Risk Minimization based models degrade drastically when training and testing distributions are different \cite{peters2016causal,arjovsky2019invariant}. To relax the i.i.d. assumption, Peters \etal~\cite{peters2016causal} suggest exploiting the ``invariance principle'' for better generalization. Unfortunately, Invariant Risk Minimization methods for deep neural networks as of yet fail to deliver competitive results, as observed in~\cite{gulrajani2020search, lin2021empirical, lin2022bayesian}. To alleviate this limitation, Lin \etal~\cite{lin2022bayesian} propose a Bayesian treatment of Invariant Risk Minimization that alleviates overfitting with deep models by defining a regularization term over the posterior distribution of classifiers, minimizing this term, and pushing the model's backbone to learn invariant features. We take inspiration from this Bayesian Invariant Risk Minimization~\cite{lin2022bayesian} and propose the first \textit{Bayesian prompt learning} approach. 

We make three contributions. First, we frame prompt learning from the Bayesian perspective and formulate it as a variational inference problem (see Figure~\ref{fig:splash}). This formulation provides several benefits. First, it naturally injects noise during prompt learning and induces a regularization term that encourages the model to learn informative prompts, scattered in the prompt space, for each downstream task. As a direct result, we regularize the prompt space, reduce overfitting to seen prompts, and improve  generalization on unseen prompts. Second, our framework models the input prompt space in a probabilistic manner, as an a priori distribution which makes our proposal compatible with prompt learning approaches that are unconditional~\cite{zhou2022learning} or conditional on the image~\cite{zhou2022conditional}. 
% \cs{Needs to be more precise.}
Third, we empirically demonstrate on 15 benchmarks that Bayesian prompt learning provides an appropriate coverage of the prompt space, prevents learning spurious features, and exploits transferable invariant features, leading to a better generalization of unseen prompts, even across different datasets and domains.

\section{Related Work}
\label{sec:related}
\vspace{-0.5em}

\textbf{Prompt learning in language.} 
Prompt learning was originally proposed within natural language processing (NLP), by models such as GPT-3~\cite{brown2020language}. Early methods constructed prompts by combining words in the language space such that the model would perform better on downstream evaluation \cite{shin2020autoprompt,jiang2020can}.
Li and Liang \cite{li2021prefix} prepend a set of learnable prompts to the different layers of a frozen model and optimize through back-propagation.
In parallel, Lester \etal \cite{lester2021power} demonstrate that with no intermediate layer prefixes or task-specific output layers, adding prefixes alone to the input of the frozen model is enough to compete with fine-tuning.
Alternatively, \cite{he2022hyperprompt} 
%focuses on a multi-task scenario and 
uses a HyperNetwork to conditionally generate task-specific and layer-specific prompts pre-pended to the values and keys inside the self-attention layers of a frozen model. Inspired by progress from NLP, we propose a  prompt learning method intended for image and language models. 

\textbf{Prompt learning in image and language.} 
Zhou \etal \cite{zhou2022learning} propose Context Optimization (CoOp), a prompt learner for CLIP, which optimizes prompts in the continuous space through back-propagation. The work demonstrates the benefit of prompt learning over prompt engineering. While \coop obtains good accuracy on prompts seen during training, it has difficulty generalizing to unseen prompts. It motivated Zhou \etal to introduce Conditional Context Optimization (CoCoOp) \cite{zhou2022conditional}. It generates instance-specific prompt residuals through a conditioning mechanism dependent on the image data, which generalizes better.% than CoOp. 
ProGrad by Zhu \etal~\cite{zhu2022prompt} also strives to bridge the generalization gap by matching the gradient of the prompt to the general knowledge of the CLIP model to prevent forgetting. 
Alternative directions consist of test-time prompt learning~\cite{testtimept_neurips22}, where consistency across multiple views is the supervisory signal, and unsupervised prompt learning~\cite{unsup_prompt22}, where a pseudo-labeling strategy drives the prompt learning.  More similar to ours is ProDA by Lu \etal~\cite{lu2022prompt}, who propose an ensemble of a fixed number of hand-crafted prompts and model the distribution exclusively within the language embedding. Conversely, we prefer to model the input prompt space rather than relying on a fixed number of templates, as it provides us with a mechanism to cover the prompt space by sampling. Moreover, our approach is not limited to unconditional prompt learning from the language embedding, but like Zhou \etal \cite{zhou2022conditional}, also allows for prompt learning conditioned on an image. 

\textbf{Prompt learning in vision and language.}  While beyond our current scope, it is worth noting that prompt learning has been applied to a wider range of vision problems and scenarios, which highlights its power and flexibility. Among them are important topics such as unsupervised domain adaptation~\cite{ge2022domain}, multi-label classification~\cite{sun2022dualcoop}, video classification~\cite{ju2021prompting}, object detection~\cite{detpro_cvpr22,promptdet_eccv22} and pixel-level labelling~\cite{rao2022denseclip}. Finally, prompt learning has also been applied to vision only models~\cite{jia2022visual,sandler2022fine} providing an efficient and flexible means to adapt pre-trained models.

\textbf{Variational inference in computer vision.} 
Variational inference  and, more specifically, variational autoencoder variants have been extensively applied to computer vision tasks as diverse as image generation \cite{ramesh2021zero, saharia2022photorealistic, rombach2022high}, action recognition~\cite{mehrasavariational}, 
%layout generation~\cite{arroyo2021variational}, 
instance segmentation~\cite{homayounfar2020levelset}, 
%pedestrian detection~\cite{zhang2021variational}, 
%super resolution~\cite{hyun2020varsr}, 
few-shot learning~\cite{zhang2019variational, schonfeld2019generalized}, 
%domain adaptation~\cite{takahashi2020partially}, 
domain generalization~\cite{du2020learning}, and continual learning~\cite{derakhshani2021kernel}. 
For example, Zhang \etal \cite{zhang2019variational} focus on the reduction of noise vulnerability and estimation bias in few-shot learning through variational inference. In the same vein, Du \etal \cite{du2020learning} propose a variational information bottleneck to better manage prediction uncertainty and unknown domains. 
Our proposed method also shares the advantages of variational inference in avoiding overfitting in low-shot settings, improving generalization, and encourages the prompt space to be resilient against these challenges. To the best of our knowledge we are the first to introduce variational inference in prompt learning.

\section{Method}
\vspace{-0.5em}

\subsection{Background}

\textbf{Contrastive Language-Image Pretraining (CLIP)}~\cite{clip_icml21} consists of an \textit{image encoder} $f(\x)$ and \textit{text encoder} $g(\txt)$, each producing a $d$-dimensional ($L_2$ normalized) embedding from an arbitrary image $\x \in \mathbb{R}^{3 \times H \times W}$, and word embeddings $\txt \in \mathbb{R}^{L \times e}$, with $L$ representing the text length and $e$ the embedding dimension\footnote{In CLIP the word embedding is learned together with the text encoder. A \textit{tokenizer} is used to convert the text into one-hot vectors, or tokens, that can be directly mapped into the word embeddings. For the sake of clarity we refer indistinctly to words and word embeddings.}. Both encoders are trained together using a contrastive loss from a large-scale dataset composed of paired images and captions. Once trained, CLIP enables zero-shot $C$-class image classification by generating each of the $c$ classifier weights $\w_c$ as the $d$-dimensional text encoding $g(\txt_c)$. Here $\txt_c$ results from adding the class-specific word embedding $\e_c$ to a pre-defined prompt $\p \in \mathbb{R}^{L-1 \times e}$, \ie, $\w_c {=} g(\txt_c)$ with $\txt_c {=} \{\p,\e_c\}$. The prompt $\p$ is manually crafted to capture the semantic meaning of the downstream task, \eg, $\txt_c {=} ``\texttt{An image of a \{class\}}"$. The probability of image $\x$ being classified as $y \in \{1...C\}$ is thus defined as 
$p(y|\x) {=} \frac{e^{f(\x)^T\w_y}}{\sum_{c}^C e^{f(\x)^T \w_{c}}}$. 
%$p(y|\x) {=} \frac{e^{f(\x)^T\w_y}}{\sum_{c'=1}^C e^{f(\x)^T \w_{c'}}}$. 

% ---- CoOp
\textbf{Context Optimization (CoOp)}~\cite{zhou2022learning} provides a learned alternative to manually defining prompts. CoOp learns a fixed prompt from a few annotated samples. The prompt is designed as a learnable embedding matrix $\p \in \mathbb{R}^{L \times e}$ which is updated via back-propagating the classification error through the frozen CLIP model. Specifically, for a set of $N$ annotated meta-training samples $\{\x_i, y_i\}_{i=1}^N$, the prompt $\p$ is obtained by minimizing the cross-entropy loss, as:
\begin{equation}
\label{eq:coop}
   \p^* = \arg \min_\p \mathbb{E}_{\x_i,y_i} [-\log p(y_i | \x_i, \p)].
\end{equation}
Note that this approach, while resembling that of common meta-learning approaches, can still be deployed in a zero-shot scenario provided that for new classes the classification weights will be given by the text encoder. Although this approach generalizes to new tasks with few training iterations, learning a fixed prompt is sensitive to domain shifts between the annotated samples and the unseen prompts.

\textbf{Conditional Prompt Learning (CoCoOp)}~\cite{zhou2022conditional} attempts to overcome domain shifts by learning an instance-specific continuous prompt that is conditioned on the input image. To ease the training of a conditional prompt generator, \cocoop defines each conditional token in a residual way, with a task-specific, learnable set of tokens $\p$ and a residual prompt that is conditioned on the input image. Assuming $\p$ to be composed of $L$ learnable tokens $\p {=} [\p_1, \p_2, \cdots , \p_L]$, the residual prompt $\rvr(\x) {=} \pi_\phi(f(\x)) \in \mathbb{R}^e$ is produced by a small neural network $\pi_\phi$ with as input the image features $f(\x)$. The new prompt is then computed as $\p(\x) {=} [\p_1 + \rvr(\x), \p_2 + \rvr(\x) , \cdots , \p_L + \rvr(\x)]$. The training now comprises learning the task-specific prompt $\p$ and the parameters $\phi$ of the neural network $\pi_\phi$. Defining the context-specific text embedding $\txt_c(\x) {=} \{\p(\x), \e_c \}$, and $p(y | \x)$ as :
\begin{equation}
\label{eq:prob_y}
    p(y | \x) = \frac{ e^{ f(\x)^T g( \txt_c(\x) )} }{ \sum_{c}^C e^{ f(\x)^T g( \txt_{c}(\x) ) } },
\end{equation}
the learning is formulated as:
\begin{equation}
    \label{eq:cocoop}
   \p^*, \phi^* = \arg \min_{\p, \phi} \mathbb{E}_{\x_i,y_i} [-\log p(y_i | \x_i, \p, \phi)].
\end{equation}
While \cocoop achieves good results in many downstream tasks, it is still prone to the domain shift problem, considering that $\pi_\phi$ provides a deterministic residual prompt from the domain-specific image features $f(\x)$.% which are expected to be domain-specific.

% -------------- ProDA
\textbf{Prompt Distribution Learning (ProDA)}~\cite{lu2022prompt} learns a distribution of prompts that generalize to a broader set of tasks. It learns a collection of prompts $\mathbf{P}{=}\{\p_k\}_{k=1}^\mathcal{K}$ that subsequently generate an \textit{a posteriori} distribution of the classifier weights for each of the target classes. For a given mini-batch of $K$ sampled prompts $\p_k {\sim} \mathbf{P}$, the classifier weights $\w_c$ are sampled from the posterior distribution $\mathbf{q}{=}\mathcal{N}(\mu_{\w_{1:C}}, \Sigma_{\w_{1:C}})$, with mean $\mu_{\w_{1:C}}$ and covariance $\Sigma_{\w_{1:C}}$ computed from the collection $\{\w_{k,c} {=} g(\txt_{k,c})\}_{c=1:C, k=1:K}$, with $\txt_{k,c} = \{\p_k, \e_c\}$. The objective is formulated as: 
\begin{equation}
\label{eq:variational_coop}
   \mathbf{P}^* = \arg \min_{\mathbf{P}} \mathbb{E}_{\x_i,y_i} [-\log \mathbb{E}_{\w_l \sim \mathbf{q}}\, p(y_i | \x_i, \w_l)].
\end{equation}
Computing $\mathbb{E}_{\w_l} \, p(y_i | \x_i, \w_l)]$ is intractable and an upper bound to Eq.~\ref{eq:variational_coop} is derived. During inference, the classifier weights are set to those given by the predictive mean $\w_c {=} \mu_{\w_{1:C}}$, computed across the set of 
learned prompts $\mathbf{P}$. 

% --------------------------------------------

\subsection{Conditional Bayesian Prompt Learning}
\label{ssec:method_conditional}

We propose to model the input prompt space in a probabilistic manner, as an \textit{a priori}, conditional distribution. We define a distribution $p_\gamma$ over the prompts $\p$ that is conditional on the image, \ie, $\p \sim p_\gamma(\x)$. To this end, we assume that $\p$ can be split into a fixed set of prompts $\p_i$ and an conditional residual prompt $\rvr$ that act as a latent variable over $\p$. The conditional prompt is then defined as:
\begin{equation}
\label{eq:new_prompt}\resizebox{.9\linewidth}{!}{$
\p_\gamma(\x) = [\p_1 + \rvr_\gamma, \p_2 + \rvr_\gamma, \cdots, \p_L + \rvr_\gamma], \, \, \rvr_\gamma \sim p_\gamma(\x),$}  % \Sigma(\x))
\end{equation}
 where $p_\gamma(\x)$ refers to the real posterior distribution over $\rvr$ conditioned on the observed features $\x$. Denoting the class-specific input as $\txt_{c,\gamma}(\x) {=} \{\p_\gamma(\x), \e_c\}$, the marginal likelihood $p(y | \x)$ is:
\begin{equation}
     \label{eq:probabilistic}
 p(y | \x) = \int_\gamma \frac{ e^{ f(\x)^T g( \txt_{c,\gamma}(\x) )} }{ \sum_{c'} e^{ f(\x)^T g( \txt_{c',\gamma}(\x) ) } } p(\p_\gamma(\x)) d \gamma.
\end{equation}
Solving Eq.~\ref{eq:cocoop} with the marginal likelihood from Eq.~\ref{eq:probabilistic} is intractable, as it requires computing $p_\gamma(\rvr | \x) p_\gamma(\x)$. Instead, we resort to deriving a lower bound by introducing a variational posterior distribution $\pi_\phi(\x)$ from which the residual $\rvr_\gamma$ can be sampled. The variational bound is defined as:
\begin{equation}
\label{eq:elbo}
\resizebox{.9\linewidth}{!}{$\log p(y | \x) \geq \mathbb{E}_{\pi_\phi(\rvr | \x)} [\log p(y | \x, \rvr)] - D_{\mathrm{KL}}\big[\pi_{\phi}(\mathbf{r} | \mathbf{z}) \| p_{\gamma}(\mathbf{r})\big],$}
\end{equation}
with $p(y | \x, \rvr) {\propto} e^{ f(\x)^T g( \txt_{c,\gamma}(\x) )}$, where the dependency on $\rvr$ comes through the definition of $\txt_{c,\gamma}$.  Following standard variational optimization practices~\cite{kingma2013auto,gordon2018metalearning}, we define $\pi_\phi$ as a Gaussian distribution conditioned on the input image features $\x$, as $\rvr(\x) {\sim} \mathcal{N}( \mu(\x), \Sigma(\x ))$, with $\mu$ and $\Sigma$ parameterized by two linear layers followed by two linear heads on top to estimate the $\mu$ and $\Sigma$ of the residual distribution. The prior $p_\gamma(\rvr)$ is defined as $\mathcal{N}(\mathbf{0},\mathbf{I})$, and we use the reparameterization trick to generate Monte-Carlo samples from $\pi_\phi$ to maximize the right side of Eq.~\ref{eq:elbo}. 
The optimization of Eq.~\ref{eq:elbo} comprises learning the prompt embeddings $\{\p_i\}_{i=1}^L$ as well as the parameters of the posterior network $\pi_\phi$ and the linear layers parameterizing $\mu$ and $\Sigma$. Note that this adds little complexity as it requires learning $\p$ and $\pi_\phi$, given that $\mu$ and $\Sigma$ are defined as two linear layers on top of $\pi_\phi$.

\textbf{Inference.} At test time, $K$ residuals are sampled from the conditional distribution $\pi_\phi(\x)$, which are used to generate $K$ different prompts per class $\p_{k} {=} [\p_1 + \rvr_k, \p_2 + \rvr_k, \cdots, \p_L + \rvr_k]$. Each prompt is prepended to the class-specific embedding to generate a series of $K$ separate classifier weights $\w_{k,c}$. We then compute $p(y {=} c \,| \, \x) {=} (1/K) \sum_{k=1}^K p(y {=} c \, | \, \x, \w_{k,c})$ and select $\hat{c} {=} \arg \max_{c} p(y {=} c \, | \, \x)$ as the predicted class. It is worth noting that because the posterior distribution is generated by the text encoder, it is not expected that for $K \rightarrow \infty$, $(1/K) \sum_{k} p( y {=} c \, | \, \x, \w_{c,k}) \rightarrow p( y {=} c | \x, g(\{\mu(\x), \e_c\})$, meaning that sampling at inference time remains relevant. 
% We study the dependency on the number of samples in the ablations.

\subsection{Unconditional Bayesian Prompt Learning}
\label{ssec:method_unconditional}
Notably, our framework can also be reformulated as an unconditional case  by simply removing the dependency of the input image from the latent distribution. In such a scenario, we keep a fixed set of prompt embeddings and learn a global latent distribution $p_\gamma$ over the residual prompts $\rvr$, as $\rvr \sim \mathcal{N}( \mu, \Sigma)$, where $\mu$ and $\Sigma$ are parameterized by two learnable vectors. To this end, we assume that $\p$ can be split into a fixed set of prompts $\p_i$ and a residual prompts $\rvr$ that act as a latent variable over $\p$. For the unconditional case, the prompt is defined as:
\begin{equation}
    \label{eq:unconditional_prompt}
\p_\gamma = [\p_1 + \rvr_\gamma, \p_2 + \rvr_\gamma, \cdots, \p_L + \rvr_\gamma], \, \, \rvr_\gamma \sim p_\gamma.
\end{equation}
In this case, $p_\gamma$ is a general distribution learned during training with no dependency on the input sample $\x$. 

Having defined both \textit{unconditional Bayesian prompt learning} and \textit{conditional Bayesian prompt learning} we are now ready to evaluate their generalization ability.
\section{Experiments and Results}
\vspace{-0.5em}
\label{sec:experiments}

\subsection{Experimental Setup}

\textbf{Three tasks and fifteen datasets.} We consider three tasks: \textit{unseen prompts generalization}, \textit{cross-dataset prompts generalization}, and \textit{cross-domain prompts generalization}. For the first two tasks, we rely on the same 11 image recognition datasets as Zhou \etal~\cite{zhou2022learning,zhou2022conditional}. These include image classification (ImageNet~\cite{deng2009imagenet} and Caltech101~\cite{fei2004learning}), fine-grained classification  (OxfordPets~\cite{parkhi2012cats}, StanfordCars~\cite{krause20133d}, Flowers102~\cite{nilsback2008automated}, Food101~\cite{bossard2014food} and FGVCAircraft~\cite{maji2013fine}), scene recognition (SUN397~\cite{xiao2010sun}), action recognition (UCF101~\cite{soomro2012dataset}), texture classification (DTD~\cite{cimpoi2014describing}), and satellite imagery recognition (EuroSAT~\cite{helber2019eurosat}). 
For cross-domain prompts generalization, we train on  ImageNet and report on ImageNetV2~\cite{recht2019imagenet}, ImageNet-Sketch~\cite{wang2019learning}, ImageNet-A~\cite{hendrycks2021natural}, and ImageNet-R~\cite{hendrycks2021many}.

\textbf{Evaluation metrics.} For all three tasks we report average accuracy and standard deviation.

\textbf{Implementation details.}
Our conditional variational prompt learning contains three sub-networks: an image encoder $f(\x)$, a text encoder $g(\txt)$, and a posterior network $\pi_{\phi}$. The image encoder $f(\x)$ is a ViT-B/16 \cite{dosovitskiy2020image} and the text encoder $g(\txt)$ a transformer \cite{vaswani2017attention}, which are both initialized with CLIP's pre-trained weights and kept frozen during training, as in \cite{zhou2022learning, zhou2022conditional}. 
The 
% \cs{metanet} 
posterior network
$\pi_{\phi}$ consists of two linear layers followed by an ELU activation function as trunk and two linear heads on top to estimate the $\mu$ and $\Sigma$ of the residual distribution. For each task and dataset, we optimize the number of samples $K$ and epochs. Other hyper-parameters as well as the training pipeline in terms of few-shot task definitions are identical to \cite{zhou2022learning, zhou2022conditional} (see Table 9 and 10 in the appendix). 
\subsection{Comparisons}

We first compare against  
\coop \cite{zhou2022learning}, \cocoop \cite{zhou2022conditional}, and ProDA~\cite{lu2022prompt} in terms of the generalization of learned prompts on unseen classes, datasets, or domains. 
For CoOp and CoCoOp, all results are adopted from~\cite{zhou2022conditional}, and we report results for ProDA using our re-implementation. 

\textbf{Task I: unseen prompts generalization} 
We report the unseen prompts generalization of our method on $11$ datasets for three different random seeds. Each dataset is divided into two disjoint subsets: seen classes and unseen classes. We train our method on the seen classes and evaluate it on the unseen classes. For a fair comparison, we follow \cite{zhou2022learning, zhou2022conditional} in terms of dataset split and number of shots. %(see \cs{Table 9 and Table 10} in the appendix.)
From Table~\ref{tab:base-to-new}, it can be seen that our best-performing method, conditional Bayesian prompt learning, outperforms \coop and \cocoop in terms of unseen prompts generalization by $11.72\%$ and $3.25\%$, respectively. 
Our proposal also demonstrates minimal average variance across all datasets. This is achieved by regularizing the optimization by virtue of the variational formulation. Moreover, our model also performs better than ProDA~\cite{lu2022prompt}, its probabilistic counterpart, by $2.64\%$. This mainly happens because our proposed method learns the prompt distribution directly in the prompt space and allows us to sample more informative prompts for the downstream tasks.

\begin{table}[t!]
\centering
\caption{\textbf{Task I: unseen prompts generalization} comparison between conditional Bayesian prompt learning and alternatives. 
% We average our accuracy over three random seeds. 
Our model provides better generalization on unseen prompts compared to CoOp, \cocoop and ProDA.}
\vspace{-2mm}
\resizebox{1\linewidth}{!}{
\begin{tabular}{lcccb}
\toprule
 & 
\multicolumn{1}{c}{\centering \textbf{CoOp}  \cite{zhou2022learning}} & 
\multicolumn{1}{c}{\centering \textbf{CoCoOp}  \cite{zhou2022conditional}} & 
\multicolumn{1}{c}{\centering \textbf{ProDA}  \cite{lu2022prompt}} & 
\multicolumn{1}{b}{\centering \textbf{Ours}}
\\ \midrule
Caltech101    & 89.81 & 93.81           & 93.23  & \textbf{94.93}\scriptsize{$\pm$0.1}  \\
DTD           & 41.18 & 56.00           & 56.48  & \textbf{60.80}\scriptsize{$\pm$0.5}\\
EuroSAT       & 54.74 & 60.04           & 66.00  & \textbf{75.30}\scriptsize{$\pm$0.7}\\
FGVCAircraft  & 22.30 & 23.71           & 34.13  & \textbf{35.00}\scriptsize{$\pm$0.5}\\
Flowers102    & 59.67 & \textbf{71.75}  & 68.68  & \textbf{70.4}0\scriptsize{$\pm$1.8} \\
Food101       & 82.26 & 91.29           & 88.57  & \textbf{92.13}\scriptsize{$\pm$0.1}\\
ImageNet      & 67.88 & 70.43           & 70.23  & \textbf{70.93}\scriptsize{$\pm$0.1}  \\
OxfordPets    & 95.29 & 97.69           & 97.83  & \textbf{98.00}\scriptsize{$\pm$0.1}  \\
StanfordCars  & 60.40 & \textbf{73.59}  & 71.20  & 73.23\scriptsize{$\pm$0.2}\\
SUN397        & 65.89 & 76.86           & 76.93  & \textbf{77.87}\scriptsize{$\pm$0.5}\\
UCF101        & 56.05 & 73.45           & 71.97  & \textbf{75.77}\scriptsize{$\pm$0.1}\\
\midrule
\textit{Average }   & 63.22 & 71.69 & 72.30 & \textbf{74.94}\scriptsize{$\pm$0.2}  \\
\bottomrule
\end{tabular}%
}
\label{tab:base-to-new}
\vspace{-5mm}
\end{table}

\textbf{Task II: cross-dataset prompts generalization}
For the next task, cross-dataset prompts generalization, the model is trained on a source dataset (ImageNet) and then assessed on 10 distinct target datasets. This experiment tries to determine how effectively our method generalizes beyond the scope of a single dataset. 
As reported in Table~\ref{tab:cross-and-domain}, our conditional Bayesian prompt learning outperforms \coop and \cocoop on the target dataset  by $2.07\%$ and $0.36\%$. This highlights that our method encourages the model to exploit transferable invariant features beneficial for datasets with a non-overlapping label space. Furthermore, our method improves accuracy for 7 out of 10 target datasets. Unlike CoCoOp, which performs better in ImageNet-like datasets such as Caltech101 and StanfordCars, our Bayesian method exhibits improvement on dissimilar datasets (\eg, FGVCAircraft, DTD, and EuroSAT), demonstrating its capacity to capture the unique characteristics of each dataset.

\begin{table}[t!]
\centering
\caption{\textbf{Task II: cross-dataset prompts generalization}. Our proposed model is evaluated on 11 datasets with different label spaces. As shown, conditional Bayesian prompt learning performs better than non-Bayesian alternatives on 7 out of 10 datasets. \textbf{Task III: cross-domain prompts generalization}. Our model is evaluated on four datasets sharing the same label space as the training data. Our method outperforms alternatives on 3 out of 4 datasets.}
\vspace{-2mm}
\resizebox{1\linewidth}{!}{%
\begin{tabular}{lccb}
\toprule
 & 
\multicolumn{1}{c}{\centering \textbf{CoOp} \cite{zhou2022learning}} & 
\multicolumn{1}{c}{\centering \textbf{CoCoOp} \cite{zhou2022conditional}} & 
\multicolumn{1}{b}{\centering \textbf{Ours}}
\\ 
\midrule
\rowcolor{palegray}
\multicolumn{4}{c}{\centering \textbf{Task II: cross-dataset prompts generalization}} \\
\midrule
Caltech101              & 93.70 & \textbf{94.43} & 93.67\scriptsize{$\pm$0.2}  \\
DTD                     & 41.92 & 45.73 & \textbf{46.10}\scriptsize{$\pm$0.1}  \\
EuroSAT                 & \textbf{46.39} & 45.37 & \textbf{45.87}\scriptsize{$\pm$0.7}  \\
FGVCAircraft            & 18.47 & 22.94 & \textbf{24.93}\scriptsize{$\pm$0.2}  \\
Flowers102              & 68.71 & 70.20 & \textbf{70.90}\scriptsize{$\pm$0.1}  \\
Food101                 & 85.30 & 86.06 & \textbf{86.30}\scriptsize{$\pm$0.1}  \\
OxfordPets              & 89.14 & 90.14 & \textbf{90.63}\scriptsize{$\pm$0.1}  \\
StanfordCars            & 64.51 & \textbf{65.50} & 65.00\scriptsize{$\pm$0.1}  \\
SUN397                  & 64.15 & 67.36 & \textbf{67.47}\scriptsize{$\pm$0.1}  \\
UCF101                  & 66.55 & 68.21 & \textbf{68.67}\scriptsize{$\pm$0.2}  \\
\midrule
\textit{Average }                & 63.88 & 65.59 & \textbf{65.95}\scriptsize{$\pm$0.2}  \\
\midrule
\midrule
\rowcolor{palegray}
\multicolumn{4}{c}{\centering \textbf{Task III: cross-domain prompts generalization}} \\
\midrule
ImageNetV2              & \textbf{64.20} & 64.07 & \textbf{64.23}\scriptsize{$\pm$0.1}  \\
ImageNet-Sketch         & 47.99 & 48.75 & \textbf{49.20}\scriptsize{$\pm$0.0}  \\
ImageNet-A              & 49.71 & 50.63 & \textbf{51.33}\scriptsize{$\pm$0.1}  \\
ImageNet-R              & 75.21 & 76.18 & \textbf{77.00}\scriptsize{$\pm$0.1}  \\
\midrule
\textit{Average }       & 59.27 & 59.88 & \textbf{60.44}\scriptsize{$\pm$0.1}  \\
\bottomrule
\end{tabular}%
}
\label{tab:cross-and-domain}
% \vspace{-5mm}
\end{table}

\textbf{Task III: cross-domain prompts generalization}
Lastly, we examine our conditional Bayesian prompt learning through the lens of distribution shift and robustness. We train our model on the source dataset ImageNet for three different random seeds, and assess it on ImageNetV2, ImageNet-Sketch, ImageNet-A, and ImageNet-R. Prior works such as \coop \cite{zhou2022learning} and \cocoop \cite{zhou2022conditional} demonstrate empirically that learning a soft-prompt improves the model's resilience against distribution shift and adversarial attack. Following their experiments, we are also interested in determining if treating prompts in a Bayesian manner maintains or improves performance. 
As reported in Table~\ref{tab:cross-and-domain} compared with CoOp, our method improves the performance on ImageNet-Sketch, ImageNet-A, and ImageNet-R by $1.21\%$, $1.62\%$, and $1.79\%$. Compared with CoCoOp, our proposed method consistently enhances the model accuracy on ImageNetV2, ImageNet-Sketch, ImageNet-A, and ImageNet-R by $0.16\%$, $0.45\%$, $0.7\%$, and $0.82\%$. This is because our proposed method prevents learning spurious features, \eg, high-frequency features, and instead learns invariant ones by virtue of the Bayesian formulation.

\begin{table}[t]
\centering
\caption{\textbf{In-domain performance.} CoOp provides the best in-domain performance, but suffers from distribution shifts. Our proposal provides the best trade-off.}
\vspace{-2mm}
\resizebox{\linewidth}{!}{%
\begin{tabular}{lcccb}
\toprule
 & 
\multicolumn{1}{c}{\centering \textbf{CoOp}~\cite{zhou2022learning}} & 
\multicolumn{1}{c}{\centering \textbf{CoCoOp}~\cite{zhou2022conditional}} &
\multicolumn{1}{c}{\centering \textbf{ProDA}~\cite{lu2022prompt}} &
\multicolumn{1}{b}{\centering \textbf{Ours}}
\\ 
\midrule
\textit{Task I}                & \textbf{82.66}  & 80.47 & 81.56 & 80.10\scriptsize{$\pm$0.1}  \\
\textit{Task II (III)}                & \textbf{71.51} & 71.02 & - & 70.70\scriptsize{$\pm$0.2}  \\
\bottomrule
\end{tabular}%
}
\label{tab:in_domain}
\vspace{-1em}
\end{table}

\textbf{In-domain performance} Different from other prompt works, we focus on prompt generalization when we have distribution shift in the input space, the label space, or both. For reference, we provide the in-domain average performance in Table
~\ref{tab:in_domain}. As expected, our generalization ability comes with reduced overfitting on the in-domain setting, leading to CoOp exhibiting better in-domain performance. However, when comparing average results in Tables~\ref{tab:base-to-new}, \ref{tab:cross-and-domain} and~\ref{tab:in_domain}, our proposal brings the best trade-off and we maintain performance on the in-domain setting while delivering a consistent performance gain on the out-of-domain setting.

\begin{table}[t!]
\centering
\caption{\textbf{Effect of variational formulation.} Formulating prompt learning as variational inference improves model generalization on unseen prompts compared to a non-Bayesian baseline~\cite{zhou2022learning}, for both the unconditional and conditional setting.}
\vspace{-2mm}
\resizebox{\linewidth}{!}{%
\begin{tabular}{lccccc}
\toprule
 & \textbf{DTD} & \textbf{EuroSAT} & \textbf{FGVC} & \textbf{Flowers102}   & \textbf{UCF101} \\ 
\midrule
Baseline        &  41.18 & 54.74 & 22.30 & 59.67 & 56.05 \\
\rowcolor{palegreen}
Unconditional   &  58.70 & 71.63 & 33.80 & 75.90 & 74.63 \\
\rowcolor{palegreen}
Conditional     &  60.80 & 75.30 & 35.00 & 70.40 & 75.77 \\
\bottomrule
\end{tabular}%
}
\label{tab:ablation-variational-formulation}
\end{table} 

\begin{table}[t]
\centering
\caption{\textbf{Benefit of the posterior distribution}.  The conditional posterior distribution $\mathcal{N}( \mu(\x), \Sigma(\x ))$ outperforms the two distributions  $\mathcal{U}(\text{0},\text{1})$ and $\mathcal{N}(\text{0},\text{I})$ by a large margin for all datasets, indicating that the conditional variant captures more informative knowledge regarding the underlying distribution of prompts in the prompt space.}
\vspace{-2mm}
\resizebox{\linewidth}{!}{%
\begin{tabular}{lcccccc}
\toprule
  & \textbf{DTD} & \textbf{EuroSAT} & \textbf{FGVC} & \textbf{Flowers102}   & \textbf{UCF101} \\ \midrule
$\mathcal{U}(\text{0},\text{1})$    
& 33.20 & 54.20 & 10.50 & 45.30 & 55.70 \\
$\mathcal{N}(\text{0},\text{I})$    
& 26.60 & 50.00	& 07.70	& 36.10 & 48.80 \\
\rowcolor{palegreen}
$\mathcal{N}( \mu(\x), \Sigma(\x ))$
& 56.40 & 64.50	& 33.00	& 72.30	& 75.60 \\
\rowcolor{palegreen}
$\mathcal{N}( \mu(\x), 0)$          
& 59.80 & 59.90 & 34.10 & 73.50 & 76.50 \\
\bottomrule
\end{tabular}%
}
\label{tab:ablation-posterior}
\vspace{-5mm}
\end{table}

\subsection{Ablations}
\label{ablation}

\textbf{Effect of variational formulation.} When prompt learning is formulated as a variational inference problem, it improves model generalization by regularizing the prompt space. To demonstrate this, we first report the model performances on unseen prompts for DTD, EuroSAT, FGVCAircraft, Flowers102, and UCF101 in Table~\ref{tab:ablation-variational-formulation}. 
%In this table, baseline, method 1, and method 2 refer to CoOp, unconditional variational prompt learning, and conditional variational prompt learning. 
Regardless of whether a variational formulation is unconditional or conditional, it consistently increases unseen prompt generalization compared with the CoOp baseline~\cite{zhou2022learning}. 
%
% \cs{This is annoying, as you do not provide an answer and instead send the reader to the next ablation.}
This is achieved by sampling more informative prompts  and  better coverage of the prompt space (which we detail in a following ablation). 
Moreover, conditional variational prompt learning enhances the model performance on 4 out of 5 datasets compared to its unconditional counterpart, indicating its effectiveness in capturing the prompt distribution. This happens as the conditional formulation provides us with the potential to successfully capture fine-grained prompts by conditioning on the input image.

\textbf{Benefit of the posterior distribution $q_\phi$.}
We ablate the effectiveness of the variational posterior distribution. To do so, we consider sampling one residual prompt from the uniform distribution 
$\mathcal{U}(\text{0},\text{1})$, normal distribution $\mathcal{N}(\text{0},\text{I})$, 
normal distribution $\mathcal{N}( \mu(\x), \Sigma(\x ))$, and normal distribution $\mathcal{N}( \mu(\x), \text{0})$ and report the unseen prompts accuracy for one random seed in Table~\ref{tab:ablation-posterior}. 
%
% \cs{Swap below paragraphs? First discuss the third row, then the last row?}
we find that drawing one sample from $\mathcal{N}( \mu(\x), \Sigma(\x ))$ yields superior results compared to drawing one sample from uniform distribution $\mathcal{U}(\text{0},\text{1})$ and normal distribution $\mathcal{N}(\text{0},\text{I})$, demonstrating the informativeness of the prompt samples generated by our conditional setting.
In addition, except for the EuroSAT dataset, a sample from the normal distribution $\mathcal{N}( \mu(\x), \text{0})$ obtains the best performance in comparison with alternatives, indicating the mean of the normal distribution $\mu(\x)$ is the most effective sample. 

\begin{table}[t]
\centering
\caption{\textbf{Influence of Monte Carlo sampling} on
unseen prompts accuracy. As demonstrated, increasing the number
of Monte Carlo samples boosts performance initially but
reaches a plateau after 20 samples for all datasets.
}
\vspace{-2mm}
\resizebox{\linewidth}{!}{%
\begin{tabular}{rcccccc}
\toprule
 & \textbf{DTD} & \textbf{EuroSAT} & \textbf{FGVC} & \textbf{Flowers102} & \textbf{UCF101} \\ \midrule
 1   & 56.40   	& 64.50	& 33.00	& 72.30 & 75.60 \\
 2   & 60.00   	& 67.40	& 33.90	& 73.90 & 76.20 \\
 5   & 62.20   	& 71.00	& 34.20	& 74.00 & 76.60 \\
 10  & 61.60   	& 73.60	& 34.40	& 73.50 & 77.00 \\
 20  & 62.90   	& 74.80	& 35.00	& 74.00 & 77.10 \\
 40  & 62.60   	& 76.10	& 35.50	& 73.80 & 77.15 \\
 %60  & 63.20   	& 76.20	& 35.20	& 74.10 & 77.20 \\
 80  & 63.50   	& 76.20	& 34.70	& 74.20 & 77.20 \\

\bottomrule
\end{tabular}%
}
\label{tab:ablation-mc}

\end{table}

\begin{table}[t]
\centering
\caption{\textbf{Influence of Monte Carlo sampling} on EuroSAT. Increasing samples increases the time complexity at inference, with a good trade-off for 20 samples (see Table~\ref{tab:ablation-mc}). }
\vspace{-2mm}
\resizebox{0.92\linewidth}{!}{%
\begin{tabular}{lccccccc}
\toprule
& \multicolumn{7}{c}{\textbf{Monte Carlo Samples}}\\
\cmidrule(lr){2-8}
& 1 & 5 & 10 & 20 & 40 & 60 & 80\\
\midrule
Time (s) & 0.02 &  0.02 &  0.03 & 0.04 & 0.07 & 0.10 &  0.13 \\
\bottomrule
\end{tabular}%
}
\label{tab:ablation-time}

\end{table}

\textbf{Influence of Monte Carlo sampling.}
When approximating the log-likelihood of input data, the number of Monte Carlo samples is an important hyperparameter. Generally, a large number of samples should lead to a better approximation and unseen prompt generalization. We ablate this hyperparameter on unseen prompts by varying the number of samples from the normal distribution $\mathcal{N}( \mu(\x), \Sigma(\x ))$ to understand the informativeness of the learned variational distribution at inference time in Table~\ref{tab:ablation-mc}. Increasing the Monte Carlo samples from $1$ to $20$  consistently improves the unseen prompts accuracy, afterwards, the accuracy saturates. 
Table \ref{tab:ablation-time} further shows the impact of the time complexity as the number of Monte Carlo samples increases on one
NVIDIA RTX 2080ti GPU. As expected, it shows the increase in sample size is accompanied by an increase in model inference time. We obtain a good trade-off in terms of accuracy and inference cost for 20 samples. For better accuracy we recommend evaluating on a larger number of Monte Carlo samples as they capture the prompt space more appropriately. We optimize this hyper-parameter using cross-validation for all datasets and use them in the remaining experiments (see Table 10 in the appendix.)

%%%%%%%%%%%%%%%%%%%%%%%%%%%%%%%%%%%%%%%%%%%%%%%%%%%%%%%%%%%%%%%%

\textbf{Effect of prompt initialization.}
Next we ablate the effectiveness of the prompt initialization on the unseen prompts accuracy for one random seed. We consider two variants. In the first we initialize the context tokens randomly using a normal distribution and in the second we initialize the context tokens with ``An image of a $\{\texttt{class}\}$''. Comparing the two variants in Table~\ref{tab:ablation-prompt-init} demonstrates that an appropriately initialized prompt consistently outperforms a randomly initialized prompt. This is mainly because the prompt samples distributed in the vicinity of a guided prompt, \eg, ``An image of a $\{\texttt{class}\}$'', are more informative than a random prompt. It can be viewed as informative prior knowledge about the prompt distribution, which facilitates the optimization towards a better prompt sub-space. 

\begin{table}[t!]
\centering
\caption{\textbf{Effect of prompt initialization.} Initializing the context tokens with an appropriate prompt ``An image of a $\{\texttt{class}\}$'' (denoted as Guided) improves the performance compared to random initialization.}
\vspace{-2mm}
\resizebox{\linewidth}{!}{%
\begin{tabular}{lccccc}
\toprule
 & \textbf{DTD} & \textbf{EuroSAT} & \textbf{FGVC} & \textbf{Flowers102} & \textbf{UCF101} \\ 
\midrule
Random     &  74.10 & 75.10 & 34.30 & 67.23 & 81.30 \\
Guided     &  74.80 & 77.90 & 35.50 & 70.05 & 82.50 \\
\bottomrule
\end{tabular}%
}
\label{tab:ablation-prompt-init}
\vspace{-5mm}
\end{table} 

\textbf{Benefit of variational ensemble.} 
Variational prompt learning can be considered as an efficient ensemble approach as it generates several samples from the prompt space and uses them in conjunction. To demonstrate the value of our proposal over a na\"ive ensemble of prompts we compare against a modified version of CoCoOp~\cite{zhou2022conditional}. We implement this version by initializing several learnable prompts per class using $M$ hand-crafted templates provided in the CLIP codebase and fine-tune them together. 
The final text feature representation is computed as a simple average of the ensemble of prompts for each class (as in Section 3.1.4 of ~\cite{clip_icml21}). For a fair comparison, the number of ensembles, $M$, equals the number of Monte Carlo samples $K$ per dataset (see Table 10 in the appendix). 
%x
The results in Table \ref{tab:ensembling} demonstrate that adding a na\"ive ensemble of prompts on top of \cocoop does not necessarily lead to an improvement.
For instance on DTD and UCF101, \cocoop performs better by itself, while on Flowers102, EuroSAT, and FGVCAircraft, the na\"ive ensemble obtains a higher unseen prompts accuracy. 
When we compare the na\"ive ensemble with our proposal, we note our method performs better in terms of unseen prompts accuracy for 4 out of 5 datasets. The reason is that, in na\"ive ensemble, these $M$ learnable prompts likely converge into contradictory prompts, that deteriorate the final performance, due to the lack of a proper regularization term in its objective function. 
%However, in our proposed method, the KL  term in Eq.~\ref{eq:elbo} encourages the model to maximize the prompt spread, finds complementary prompts, avoids this possible collapse, and provides better generalization for ensemble learning.
Moreover, the KL divergence in Eq.~\ref{eq:elbo} serves to regulate and expand the norm of the prompt space. This allows our proposal to freely explore the prompt space, leading to complementary prompt samples. We believe the prompt samples are not contradictory as they are jointly optimized to cover inherent variability in the model predictions and maximize model performance.

\begin{table}[t]
\centering
\caption{\textbf{Benefit of variational ensemble.} 
Our conditional Bayesian prompt learning outperforms \cocoop and its na\"ive ensemble for 4 out of 5 datasets regarding unseen prompts accuracy, demonstrating that our method is effective in constructing informative prompt samples for ensemble learning in a data-driven manner.}
\vspace{-2mm}
\resizebox{\linewidth}{!}{%
\begin{tabular}{lccccc}
\toprule
 & \textbf{DTD} & \textbf{EuroSAT} & \textbf{FGVC} & \textbf{Flowers102}  & \textbf{UCF101} \\ 
\midrule
\cocoop~\cite{zhou2022conditional}                     
&  56.00 & 60.04 & 23.71 & 71.75 & 73.45 \\
Na\"ive ensemble            
&  52.17 & 66.10 & 32.63 & \textbf{72.17} & 73.90 \\
\rowcolor{palegreen}
\textbf{\textit{This paper}}                   
&  \textbf{60.80} & \textbf{75.30} & \textbf{35.00} & 70.40 & \textbf{75.77} \\
\bottomrule
\end{tabular}%
}
\label{tab:ensembling}
\vspace{-5mm}
\end{table}

\begin{figure*}[t!]
\centering
\begin{subfigure}{0.96\linewidth}
  \centering
  \includegraphics[width=1\textwidth]{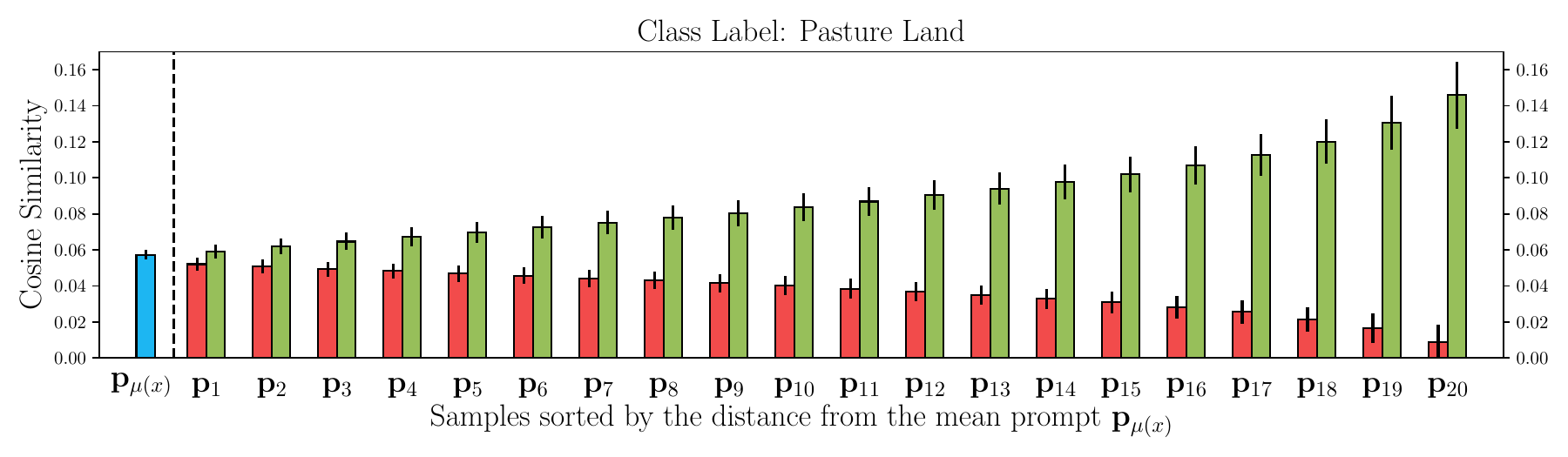}
\end{subfigure}%
\vspace{-2mm}
\caption{\textbf{Variational distribution interpretation} on the EuroSAT dataset. The text encoding of the mean prompt $\p_{\mu(\x)}$ (\sqbox{barblue}) is the most similar to the image encoding. As we move further away from the mean prompt, the cosine similarity scores between the text encoding and image encoding decrease further (\sqbox{barred}).
When we ensemble the text encoding of different prompts the cosine similarity increases (\sqbox{bargreen}), where the maximum similarity is obtained when all text encodings are combined. 
}
\vspace{-2mm}
\label{fig:interpretation}
\end{figure*}

\textbf{Variational distribution interpretation.} Next, we provide an intuition on how prompt samples are distributed within the variational distribution. For an input image $\x$ and its corresponding target $y$, $K$ residuals are sampled from the conditional distribution $\pi_\phi(\x)$, which are used to generate $K$ different prompts per class $\p_{k} {=} [\p_1 + \rvr_k, \p_2 + \rvr_k, \cdots, \p_L + \rvr_k]$. We also construct a new prompt based on the mean of the variational distribution $\mu(\x)$ dubbed as the mean prompt $\p_{\mu(\x)} {=} [\p_1 + \mu(\x), \p_2 + \mu(\x), \cdots, \p_L + \mu(\x)]$. These $K$ prompts are sorted based on their distance to the mean of the variational distribution $\mu(\x)$, where for any $i$ and $j$ ($i \leq j$), $d(\mu(\x),\rvr_i) \leq d(\mu(\x),\rvr_j)$. All these $K$ prompts and the mean prompt are prepended to the class-specific embedding to generate a series of $K$ separate text encodings $\w_{\mu(\x),c}, \w_{1,c}, \w_{2,c}, \cdots, \w_{K,c}$. For each class $y$, we compute the cosine similarity of the image encoding $f(\x)$ and $\w_{\mu(\x),y}, \w_{1,y}, \w_{2,y}, \cdots, \w_{K,y}$, and visualize them for $K{=}20$ prompts in Figure~\ref{fig:interpretation}. The mean prompt $\p_{\mu(\x)}$ has the largest cosine similarity between its corresponding text and image encoding among other prompts, indicating it is the most informative sample. As we move further away from the mean prompt, the cosine similarity score decreases. Thus, the further a prompt is from the mean prompt, the less informative it becomes. 

Next we demonstrate that prompts distributed within the variational distribution positively complement each other and maximize cosine similarity. Given the text encodings $\w_{\mu(\x),y}, \w_{1,y}, \w_{2,y}, \cdots, \w_{M,y}$ we further define a new weight as $\w_{y}^J {=} \w_{\mu(\x),y} + \sum_{j=1}^{J}{\w_{j,y}}$, where $\w_{y}^J$ is the cumulative sum of text encodings regarding prompt $\p_{1}$ to $\p_{J}$ and mean prompt $\p_{\mu(\x)}$. The cosine similarity between the image encoding $f(\x)$ and all $\{\w_{y}^j\}_{j=1}^{j=M}$ are computed and visualized in Figure~\ref{fig:interpretation}. 
As shown, summing up, or ensembling, text encodings together increases the cosine similarity, where the maximum similarity is obtained when all classifier weights are combined. Hence, our knowledge from the prompt space can be adjusted by gradually increasing the number of Monte Carlo samples.

\begin{figure}[ht]
\centering
\vspace{-2mm}
\begin{subfigure}{1\linewidth}
  \centering
  \includegraphics[width=1\linewidth,clip,trim=1cm 0cm 1cm 0cm]{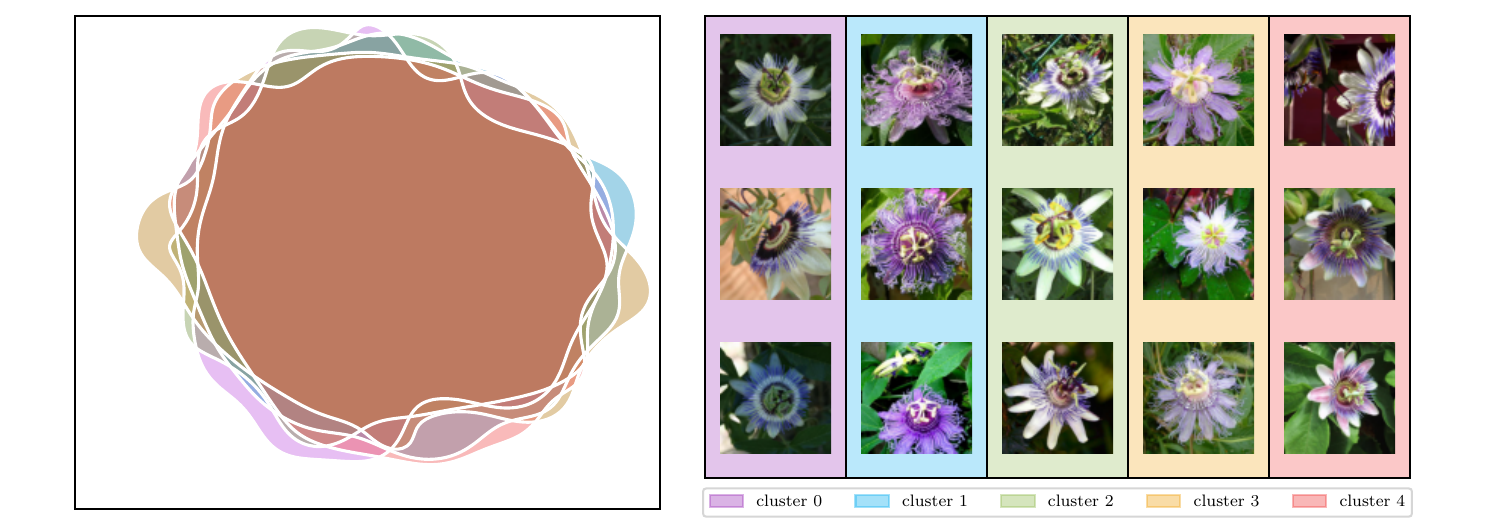}
  \caption{Passion Flower}
\end{subfigure}%

\begin{subfigure}{1\linewidth}
  \centering
  \includegraphics[width=1\linewidth,clip,trim=1cm 0cm 1cm 0cm]{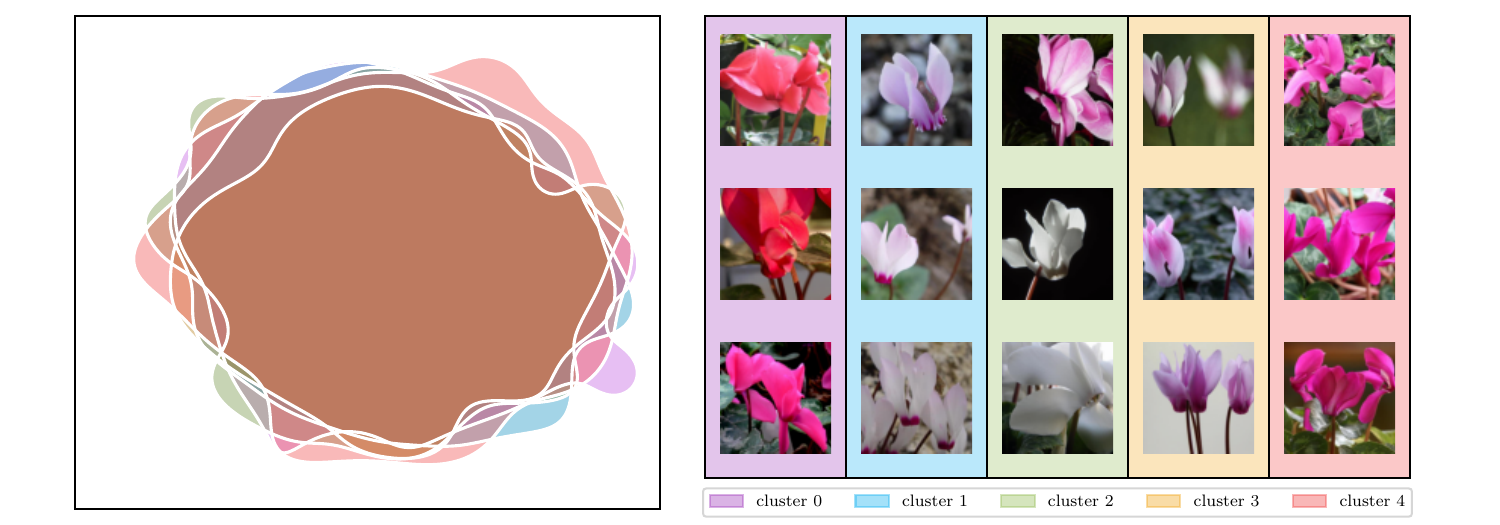}
  \caption{Cyclamen}
\end{subfigure}%

\caption{\textbf{Factor of variation analysis} on Flowers102 for two different classes. Left: we plot prompt samples across five clusters. Right: we show the top-3 most representative samples within each cluster (\eg, 3 closest images to the centroids). There is a region where the contours for five different clusters intersect, indicating shared knowledge related to its corresponding class, while they diverge slightly, indicating a particular variation factor.}
\label{fig:variation}
\vspace{-5mm}
\end{figure}

\textbf{Factor of variation analysis}
Here we analyze whether prompts that are sampled at the distributional modes correlate with features that characterize any known factors of variation (\eg, sub-domains) in a dataset. We perform this experiment on our conditional Bayesian prompt learning model.
First, we randomly select two classes
%
% $c_1$, $c_2$, and $c_3$
$c_1$ and $c_2$
from the Flowers102 dataset (Passion Flower, and Cyclamen, for reference). Then, for each class, we compute image features by forwarding the respective input images $\x$ into the image encoder of CLIP. Next, we apply $k$-means to group the image features into $Q{=}5$ cluster centroids. Note that each cluster centroid is assumed to capture some factors of visual variation within each class. Afterward, we treat the cluster assignments as pseudo-labels of those images. Note that this is done separately for each class. Additionally, for an input image $\x$, we generate $M{=}21$ different prompts. These prompts are fed into the clip's text encoder, which generates $M$ different weights $\w_{1}, \w_{2}, \cdots, \w_{M}$ per image instance. We now have a set of $M$ weights for $Q$ pseudo-labels across two classes, which are visualized in each row of Figure~\ref{fig:variation}. 
From this figure, it can be seen that there is a high intersection between the distribution of all sub-classes, representing shared knowledge. However, some particular visual differences are expressed in regions without intersections in the distribution. Hence, we believe there is a correlation between the modes of the variational distribution and visual particularities of subsets of the same class.

\section{Conclusion}
\vspace{-0.5em}

In this paper, we formulate prompt learning for image-language models as a variational inference problem. The Bayesian perspective allows regularizing the prompt space, reducing overfitting to seen prompts, and improving unseen prompt generalization. We model the input prompt space as an a priori distribution, making our framework compatible with unconditional or conditional prompt learning approaches. Using 15 benchmarks, we show that Bayesian prompt learning provides an appropriate coverage of the prompt space, prevents learning spurious features, and exploits transferable invariant features, leading to better generalization of unseen prompts, even across datasets and domains.
We believe prompt distributions can be estimated also using other probabilistic methods, like normalizing flows, energy-based models and diffusion models, 
which we will leave open for future investigation.

%------------------------------------------------------------------------

{\small
\bibliographystyle{ieee_fullname}
\bibliography{egbib}
}

\definecolor{Gray}{gray}{0.9}
\definecolor{palegray}{HTML}{F5F5F5}
\definecolor{barred}{HTML}{F24B4B}
\definecolor{barblue}{HTML}{1DB6F2}
\definecolor{baryellow}{HTML}{F2A922}
\definecolor{barpurpule}{HTML}{A441BF}
\definecolor{bargreen}{HTML}{97BF5A}
\definecolor{palepurpule}{HTML}{FFF8FF}
\definecolor{palegreen}{HTML}{F6FAF3}
\definecolor{refcolor}{HTML}{00FE2C}

\newcolumntype{b}{>{\columncolor{palegreen}}l}
\newcolumntype{q}{>{\columncolor{palegreen}}c}

\def\rvr{{\mathbf{r}}}

%%%%%%%%% TITLE
\title{Supplementary
material for Bayesian Prompt Learning for Image-Language Model Generalization}

% \and

\maketitle
\ificcvfinal\thispagestyle{empty}\fi

\begin{table}[!h]
\renewcommand\thetable{9}
\centering
\caption{Shared hyperparameters used to generate all results in the main paper.}
\vspace{-2mm}
\resizebox{\linewidth}{!}{%
\begin{tabular}{bl}
\toprule 
\textbf{Hyperparameters}   & \textbf{Values}\\
\midrule
Batch Size                 & 1              \\\midrule
Input Size                 & $224 \times 224$\\\midrule
Input Interpolation Method & ``Bicubic''\\\midrule
Input Mean                 & \begin{tabular}[c]{@{}l@{}}[0.48145466, \\    0.4578275, \\    0.40821073]\end{tabular}                     \\\midrule
Input STD                  & \begin{tabular}[c]{@{}l@{}}[0.26862954, \\    0.26130258, \\    0.27577711]\end{tabular}                    \\\midrule
Transformation             & \begin{tabular}[c]{@{}l@{}}{[}``random resized crop'', \\  ``random flip'', \\  ``normalize''{]}\end{tabular} \\\midrule
Optimizer                  & SGD          \\\midrule
Learning Rate              & $2e-3$       \\\midrule
LR Scheduler               & ``cosine''   \\\midrule
Warmup Epoch               & $1$          \\\midrule
Warmup Type                & ``Constant'' \\\midrule
Warmup LR                  & $1e-5$       \\\midrule
Backbone                   & ViT-B/16     \\\midrule
Prompt Length              & $4$          \\\midrule
Prompt Initialization      & ``a photo of a \{class\}''                                                                                    \\\midrule
Number of Shots            & $16$ \\
\bottomrule                                                                                                        
\end{tabular}
}
% \vspace{-2mm}
\label{tab:shared_hparams}
\end{table}

\section{Hyperparameters}
In this section, we provide the detailed hyperparameter settings in Tables \ref{tab:shared_hparams} and \ref{tab:dataset_hparams} that are used to generate results in the main paper for each dataset.
There are two sets of hyperparameter. In Table~\ref{tab:shared_hparams}, we report the shared hyperparameters among unconditional and conditional Bayesian prompt learning. Table~\ref{tab:dataset_hparams} contains parameters that are optimized per dataset.

\begin{table}[!h]
\centering
\renewcommand\thetable{10}
\caption{\textbf{Dataset-specific hyper-parameters} used to generate all results in the main paper. In this table, we provide the number of Monte Carlo samples (MC) and also the number of epochs used to optimize our  unconditional and conditional Bayesian prompt learning.}
\vspace{-2mm}
\resizebox{\linewidth}{!}{%
\begin{tabular}{lccccccccccc}
\toprule 
& \textbf{\rot{ImageNet}} & \textbf{\rot{Caltech101}} & \textbf{\rot{OxfordPets}} & \textbf{\rot{StanfordCars}} & \textbf{\rot{Flowers102}} & \textbf{\rot{Food101}} & \textbf{\rot{FGVC}} & \textbf{\rot{SUN397}} & \textbf{\rot{DTD}} & \textbf{\rot{EuroSAT}} & \textbf{\rot{UCF101}} \\ \midrule
MC              & 10 & 20 & 40 & 20 & 10 & 20 & 10 & 20 & 40 & 20 & 5  \\
Epochs            & 10 & 20 & 20 & 40 & 40 & 20 & 10 & 10 & 10 & 60 & 20 \\
\bottomrule
\end{tabular}
}
\vspace{-2mm}
\label{tab:dataset_hparams}
\end{table}

\begin{table}[t]
\centering
\renewcommand\thetable{11}
\caption{Comparison between conditional Bayesian prompt learning performance and CLIP model on unseen prompt generalization \textbf{(Task I)} and cross-domain prompts generalization \textbf{(Task III)}. Our model consistently performs better than CLIP model acroos all tasks.}
\vspace{-2mm}
%\resizebox{1\linewidth}{!}{%
\begin{tabular}{lcc}
\midrule
 & 
\multicolumn{1}{c}{\centering \textbf{CLIP}} & \multicolumn{1}{c}{\centering \textbf{Ours}} \\
\midrule
\rowcolor{palegray}
\multicolumn{3}{c}{\centering \textbf{Task I}} \\
\midrule
Caltech101              & 94.00 & \textbf{94.93}\scriptsize{$\pm$0.1}  \\
DTD                     & 59.90 & \textbf{60.80}\scriptsize{$\pm$0.5}  \\
EuroSAT                 & 64.05 & \textbf{75.30}\scriptsize{$\pm$0.7}  \\
FGVCAircraft            & \textbf{36.29} & 35.00\scriptsize{$\pm$0.5}  \\
Flowers102              & \textbf{77.80} & 70.40\scriptsize{$\pm$1.8}  \\
Food101                 & 91.22 & \textbf{92.13}\scriptsize{$\pm$0.1}  \\
ImageNet                & 68.14 & \textbf{70.93}\scriptsize{$\pm$0.1}  \\
OxfordPets              & 97.26 & \textbf{98.00}\scriptsize{$\pm$0.1}  \\
StanfordCars            & \textbf{74.89} & 73.23\scriptsize{$\pm$0.2}  \\
SUN397                  & 75.35 & \textbf{77.87}\scriptsize{$\pm$0.5}  \\
UCF101                  & \textbf{77.50} & 75.77\scriptsize{$\pm$0.1}  \\
\midrule
\textit{Average }       & 74.22 & \textbf{74.94}\scriptsize{$\pm$0.2}  \\
\midrule
\midrule
\rowcolor{palegray}
\multicolumn{3}{c}{\centering \textbf{Task III}} \\
\midrule
ImageNetV2              & 60.83 & \textbf{64.23}\scriptsize{$\pm$0.1}  \\
ImageNet-Sketch         & 46.15 & \textbf{49.20}\scriptsize{$\pm$0.0}  \\
ImageNet-A              & 47.77 & \textbf{51.33}\scriptsize{$\pm$0.1}  \\
ImageNet-R              & 73.96 & \textbf{77.00}\scriptsize{$\pm$0.1}  \\
\midrule
\textit{Average }       & 57.18 & \textbf{60.44}\scriptsize{$\pm$0.1}  \\
\bottomrule
\end{tabular}%

\label{tab:fixed-prompt}
% \vspace{-5mm}
\end{table}

\section{More Ablations}
\label{more_ablation}
\begin{figure*}[!h]
\renewcommand\thefigure{5}
\centering
% \begin{subfigure}{1\linewidth}
%   \centering
%   \includegraphics[width=1\textwidth]{images/iclr-backbone-base.pdf}
%   \caption{\textbf{
% Ablation of different vision encoder backbones with respect to new accuracy}. A more over-parameterized model leads to better generalization performance across all datasets.}
% \end{subfigure}%

\begin{subfigure}{1\linewidth}
  \centering
  \includegraphics[width=1\textwidth]{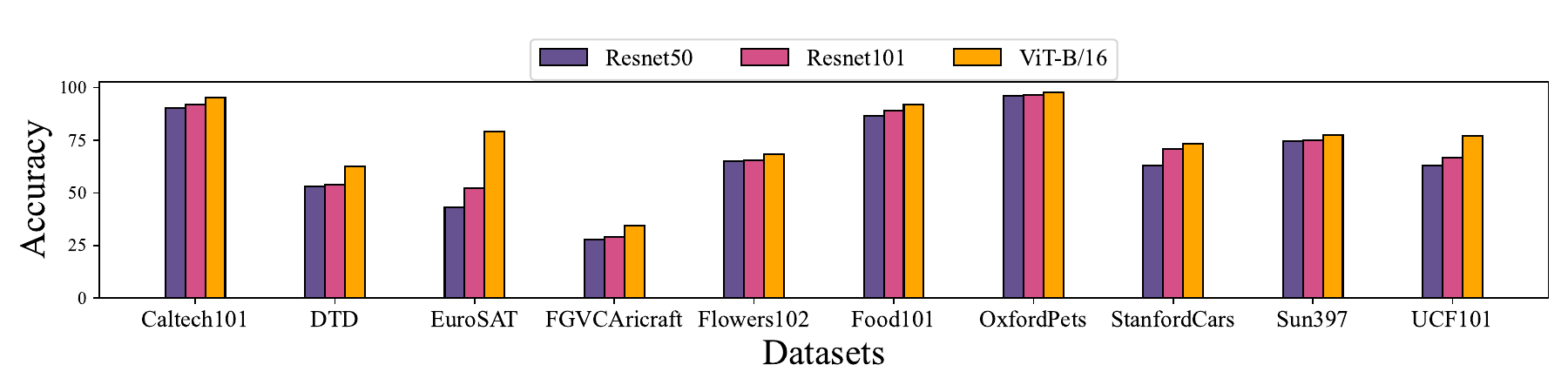}
\end{subfigure}%
\caption{\textbf{
Ablation of different vision encoder backbones with respect to unseen prompt generalization}. A more over-parameterized model leads to better generalization performance across all datasets.}
% \begin{subfigure}{1\linewidth}
%   \centering
%   \includegraphics[width=1\textwidth]{images/iclr-backbone-harmonic.pdf}
%   \caption{\textbf{
% Ablation of different vision encoder backbones with respect to harmonic mean}. A more over-parameterized model leads to better performance across all datasets.}
% \end{subfigure}%
\label{fig:variation-supp}
\end{figure*}

\textbf{Vision encoder alternatives.}
All previous experiments benefit from ViT-B/16 as the vision encoder's backbone following~[\textcolor{refcolor}{54}, \textcolor{refcolor}{55}, \textcolor{refcolor}{32}]. For completeness, in Figure~\ref{fig:variation-supp}, we replace this vision encoder with a Resnet50 and Resnet100 and examine its impact on unseen prompt generalization task for one random seed. As reported, the visual transformer outperforms the Resnet alternatives on 10 out of 10 benchmarks due to the fact that a more over-parameterized model is able to extract better generalizable features.
Hence, we suggest training and evaluating Bayesian prompt learning on visual transformer for better model performance.

\textbf{Comparison with Fixed-Prompt Baseline.} This section presents a comparison between conditional Bayesian prompt learning and the fixed-prompt baseline, such as CLIP, as shown in Table~\ref{tab:fixed-prompt}. We assess their performance in terms of unseen prompt generalization (Task I) and cross-domain prompt generalization (Task III). In the fixed-prompt approach, prompts remain non-learnable and are usually hand-engineered. In contrast, our approach involves training prompts and adapting them to downstream tasks. The experiments in Table~\ref{tab:fixed-prompt} demonstrate that our proposed method outperforms the CLIP model on 7 out of 11 datasets in Task I and on all 4 datasets in Task III. These results underscore the effectiveness of our proposed approach. 

\begin{figure*}[t!]
\centering
\begin{subfigure}{0.96\linewidth}
  \centering
  \includegraphics[width=1\textwidth]{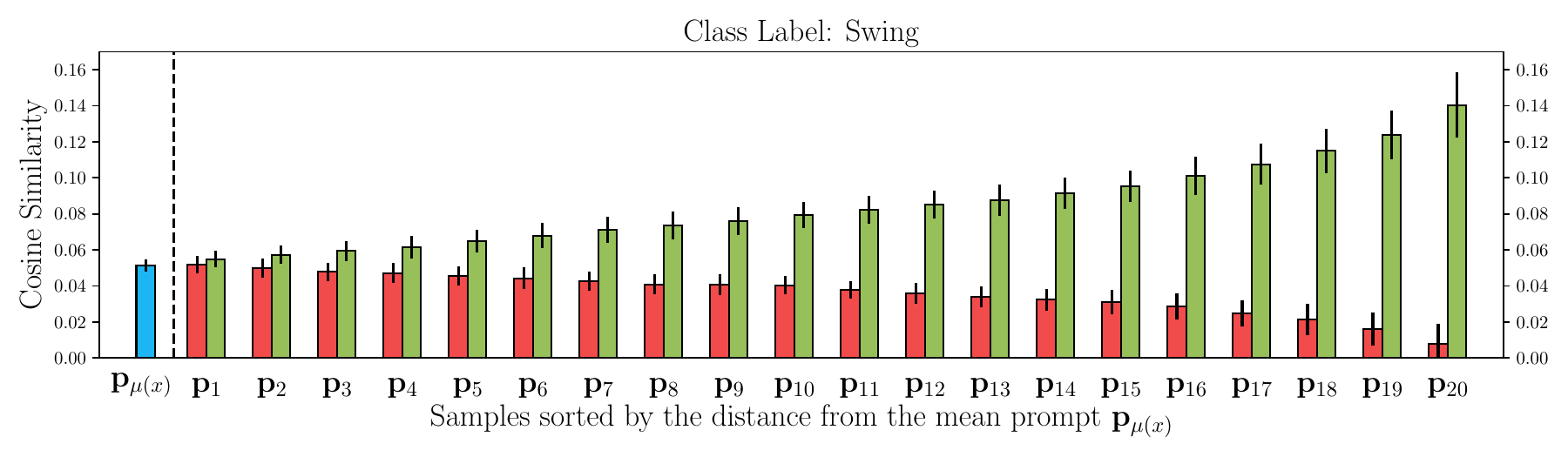}
  \includegraphics[width=1\textwidth]{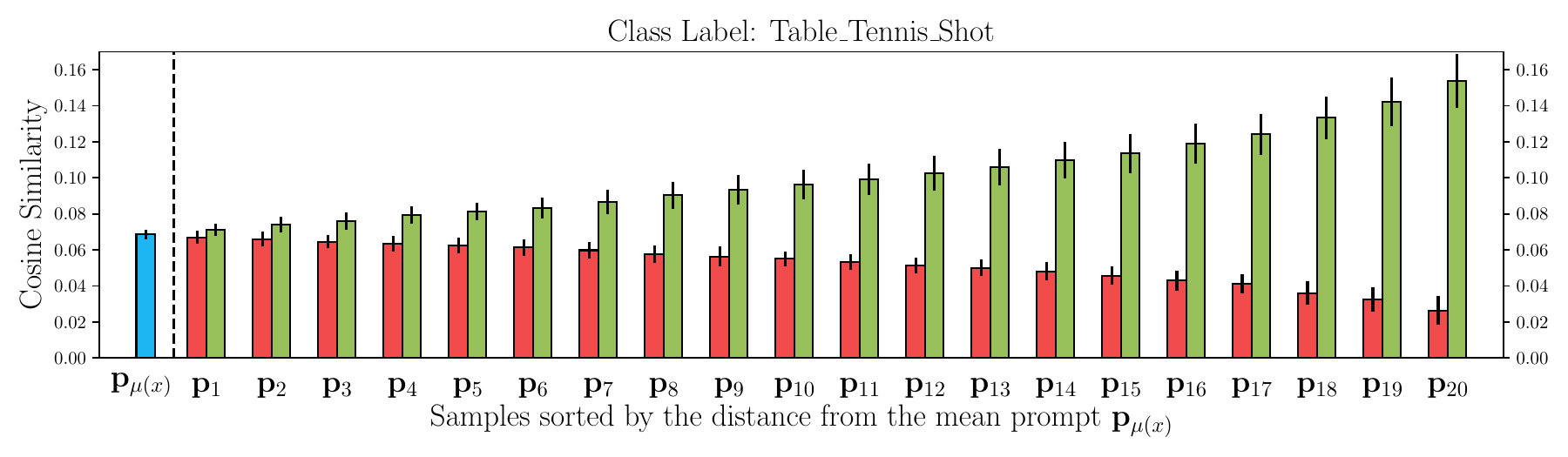}
\end{subfigure}%
\vspace{-2mm}
\caption{\textbf{Variational distribution interpretation} on the UCF101 dataset. The text encoding of the mean prompt $\p_{\mu(\x)}$ (\sqbox{barblue}) is the most similar to the image encoding. As we move further away from the mean prompt, the cosine similarity scores between the text encoding and image encoding decrease further (\sqbox{barred}).
When we ensemble the text encoding of different prompts the cosine similarity increases (\sqbox{bargreen}), where the maximum similarity is obtained when all text encodings are combined. 
}
\vspace{-2mm}
\label{fig:interpretation-2}
\end{figure*}

\end{document}